\documentclass[journal]{IEEEtran}
\usepackage[T1]{fontenc}
\usepackage{graphicx}
\usepackage{times}
\usepackage{helvet}
\usepackage{courier}
\usepackage{amsmath}
\usepackage{algorithm}
\usepackage{algorithmic}
\usepackage{csquotes} 
\usepackage{color}
\usepackage{paralist}
\usepackage{amssymb}
\usepackage{indentfirst}
\usepackage{subfigure}
\usepackage{float}
\usepackage{multirow}
\usepackage{cite}
\usepackage{mathrsfs}

\usepackage{mathrsfs} 
\usepackage{amsfonts}
\usepackage{todonotes}
\usepackage{pgfplots} 
\usepackage{epstopdf}
\usepackage{epsfig}
\pgfplotsset{compat=newest}
\usepackage{color}
\definecolor{forestgreen}{RGB}{0,139,69}
\usepackage{caption}

\usepackage{xcolor}
\definecolor{citecolor}{HTML}{0071bc}
\usepackage[colorlinks, linkcolor=red,  anchorcolor=blue, citecolor=citecolor]{hyperref} 

\usepackage{xcolor}
\definecolor{SeaGreen4}{RGB}{0,205,102} 
\definecolor{SlateBlue}{RGB}{106,90,205} 
\definecolor{DarkRed}{RGB}{178,34,34}

\usepackage[switch]{lineno}

\usepackage{textcomp,booktabs}
\usepackage{amssymb}
\usepackage{pifont}
\newcommand{\cmark}{\ding{51}}%
\newcommand{\xmark}{\ding{55}}%

\usepackage{makecell}

\usepackage{colortbl}
\definecolor{mygray}{gray}{.9}
\definecolor{mypink}{rgb}{.99,.91,.95}
\definecolor{mycyan}{cmyk}{.3,0,0,0}

\begin{document}

\title{ Retain, Blend, and Exchange: A Quality-aware Spatial-Stereo Fusion Approach for Event Stream Recognition}

\author{ Lan Chen, Dong Li, Xiao Wang*, \emph{Member, IEEE}, Pengpeng Shao, Wei Zhang, \\ 
            Yaowei Wang, \emph{Member, IEEE}, Yonghong Tian, \emph{Fellow, IEEE}, Jin Tang 
\thanks{
$\bullet$ Lan Chen is with the School of Electronic and Information Engineering, Anhui University, Hefei 230601, China. (email: chenlan@ahu.edu.cn)

$\bullet$ Dong Li, Xiao Wang, and Jin Tang are with the School of Computer Science and Technology, Anhui University, Hefei 230601, China. (email: lidongcvpr@foxmail.com, \{xiaowang, tangjin\}@ahu.edu.cn) 

$\bullet$ Pengpeng Shao is with the Department of Automation, Tsinghua University, Beijing, China. (email: ppshao@tsinghua.edu.cn)

$\bullet$ Wei Zhang is with Peng Cheng Laboratory, Shenzhen, China. (email: zhangwei1213052@126.com)   

$\bullet$ Yaowei Wang is with Peng Cheng Laboratory, Shenzhen, China; Harbin Institute of Technology (HITSZ), Shenzhen, China. (email: wangyw@pcl.ac.cn) 

$\bullet$ Yonghong Tian is with Peng Cheng Laboratory, China; National Key Laboratory for Multimedia Information Processing, School of Computer Science, Peking University, China; School of Electronic and Computer Engineering, Shenzhen Graduate School, Peking University, China (email: yhtian@pku.edu.cn) 

} 
\thanks{* Corresponding author: Xiao Wang (xiaowang@ahu.edu.cn)}  
}

\markboth{IEEE Transactions on XXX, 2024}     
{Shell \MakeLowercase{\textit{et al.}}: Bare Demo of IEEEtran.cls for IEEE Journals}

\maketitle

\begin{abstract}
Existing event stream-based pattern recognition models usually represent the event stream as the point cloud, voxel, image, etc., and design various deep neural networks to learn their features. Although considerable results can be achieved in simple cases, however, the model performance may be limited by monotonous modality expressions, sub-optimal fusion, and readout mechanisms. In this paper, we propose a novel dual-stream framework for event stream-based pattern recognition via differentiated fusion, termed EFV++. It models two common event representations simultaneously, i.e., event images and event voxels. The spatial and three-dimensional stereo information can be learned separately by utilizing Transformer and Graph Neural Network (GNN). We believe the features of each representation still contain both efficient and redundant features and a sub-optimal solution may be obtained if we directly fuse them without differentiation. Thus, we divide each feature into three levels and retain high-quality features, blend medium-quality features, and exchange low-quality features. The enhanced dual features will be fed into the fusion Transformer together with bottleneck features. In addition, we introduce a novel hybrid interaction readout mechanism to enhance the diversity of features as final representations. Extensive experiments demonstrate that our proposed framework achieves state-of-the-art performance on multiple widely used event stream-based classification datasets. Specifically, we achieve new state-of-the-art performance on the Bullying10k dataset, i.e., $90.51\%$, which exceeds the second place by $+2.21\%$. The source code of this paper has been released on \url{https://github.com/Event-AHU/EFV_event_classification/tree/EFVpp}. 
\end{abstract}


\IEEEpeerreviewmaketitle

\section{Introduction}  
\label{sec:intro} 
\IEEEPARstart{R}{ecognizing} the category of a given object is a fundamental problem in computer vision. Most of the previous classification models are developed for frame-based cameras, in other words, these recognition models focus on encoding and learning the representation of RGB frames. With the rapid development of deep learning, frame-based classification has achieved significant improvement in recent years. Representative deep models (e.g., the AlexNet~\cite{ul2018alexnet}, ResNet~\cite{he2016deep}, and Transformer~\cite{vaswani2017attention}) and benchmark datasets (e.g., ImageNet~\cite{krizhevsky2017imagenet}) are proposed one after another. However, the recognition performance in challenging scenarios is still far from satisfactory, including heavy occlusion, fast motion, and low illumination.

\begin{figure*}
    \centering
    \includegraphics[width=0.85\linewidth]{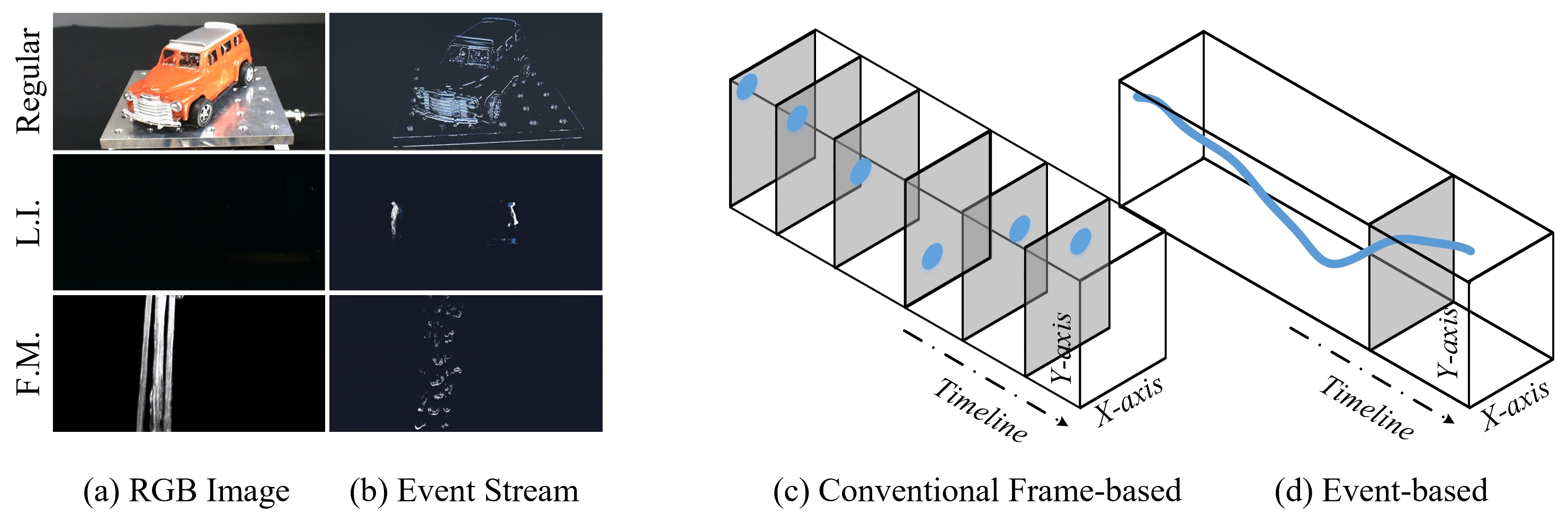}
    \caption{Comparison of the frame- and event stream-based cameras~(\url{https://youtu.be/6xOmo7Ikwzk}). 
    (a, b) shows representative samples in regular scenarios, low-illumination (L.I.), and fast motion (F.M.).  
    (c, d) illustrates the different types of raw data representation of frame- and event stream-based cameras.} 
    \label{fig:firstIMG} 
\end{figure*}

To improve object recognition in challenging scenarios, some researchers have started leveraging other sensors to obtain more effective signal inputs, thus enhancing recognition performance~\cite{sun2022human}. Among them, one of the most representative sensors is the event camera, also known as DVS (Dynamic Vision Sensor), which has been widely exploited in computer vision~\cite{tang2022revisiting, ding2024video, wang2023visevent, zhu2022eventreconstruction, chen2022residualdeblur, ding2023emlb, jiang2023eventEnhance, xie2024eisnet}. This paper focuses on using event cameras for object recognition. As shown in Fig.~\ref{fig:firstIMG}, different from the frame-based camera which records the light intensity for each pixel simultaneously, the event camera captures pulse signals asynchronously based on changes in light intensity, recording binary digital values of either zero or one. Typically, an increase in brightness is denoted as an ON event, while a decrease corresponds to an OFF event. An event pulse signal can be represented as a quadruple ($x, y, t, p$), where $x, y$ represents the spatial position information, $t$ denotes the timestamp, and $p$ represents the polarity, i.e., ON/OFF event. Many works~\cite{gallego2020eventsurvey} demonstrate that the event camera performs better in High Dynamic Range (HDR), high temporal resolution, low latency response, and strong robustness. Due to the features of spatially sparse, some human-centric tasks like pedestrian attribute recognition~\cite{wang2022PARSurvey}, action recognition~\cite{wang2022hardvs}, and re-identification~\cite{wang2021event, shu2021LAST}, can well protect human privacy. Therefore, utilizing event cameras for object recognition is a research direction that holds great research value and practical potential.

Recently, researchers have already conducted studies on object recognition using event cameras and have proposed various approaches to address this task, including CNN (Convolutional Neural Network)~\cite{wang2019ev}, GNN (Graph Neural Network)~\cite{wang2021event}, Transformer~\cite{vaswani2017attention}, etc. Although these methods have achieved good accuracy by representing and learning events from different perspectives, they are still limited by the following aspects: 
\begin{itemize}
    \item \textbf{Monotonic Event Representation}: They rely on a single event representation form, such as images, point clouds, or voxels, which may limit the expressiveness and versatility of the learned features. Different event representation forms may capture different aspects of the data, and using only one representation may lead to the loss of valuable information. 

    \item \textbf{Independent Network Structure}: The current methods are constrained to using only one of the deep learning architectures, such as CNNs, GNNs, or Transformers, for feature learning. Each architecture has its strengths and limitations in capturing different types of patterns and dependencies in data. By restricting the choice to a single architecture, the methods may not fully exploit the potential benefits and complementary strengths of different architectures. 
    
    \item \textbf{Ignore the Readout Mechanism}: Existing models usually fuse the multi-view/-modality features by concatenating them into a unified representation, however, they seldom consider mining the combinatorial relations. 
\end{itemize}
To address these limitations, we attempt to explore approaches that can integrate multiple event representation forms and leverage the combined power of different deep learning architectures. This could involve developing novel fusion techniques or hybrid architectures that can effectively capture and leverage diverse features and dependencies present in event data. By doing so, we can potentially enhance the performance and flexibility of event stream-based object recognition methods.

Based on the aforementioned observations and reflections, in this work, we propose an effective dual-stream event information processing framework, referred to as EFV++. Specifically, we first transform the dense event stream into event image and event voxel representations. For the input of image frames, we utilize advanced spatiotemporal Transformer networks to learn spatiotemporal features. For voxel input, considering the sparsity of events, we employ a top-$k$ selection method to sample meaningful signals for constructing a structured graph, and then use GNN (Graph Neural Network) to learn these volumetric structured features. More importantly, we consider the different qualities of event features in each stream and fuse them in a differentiable way. Because the two branches may contain both effective and redundant features, the key to achieving effective multi-modal fusion lies in discarding redundant features, retaining effective features, and organically integrating them. Then, we concatenate and feed the bottleneck features and the enhanced features of the dual branches into the fusion Transformer. In this paper, we also explore the impact of different feature permutations on the results and find that even better results can be obtained with features summarized by GRU (Gated Recurrent Unit) layers. An overview of our proposed EFV++ can be found in Fig.~\ref{fig:framework}.

To sum up, we conclude the main contributions of this paper as the following three aspects: 

1). Based on the dual-branch representation of events, we propose a novel multi-view discriminative interaction fusion framework, termed EVF++. This framework adeptly adapts processing based on the quality of features, selectively retaining, fusing, and exchanging information between different views. 

2). We propose a new readout mechanism based on the GRU network which fuses the features of event image, event voxel, and bottleneck features organically. 

3). Extensive experiments on multiple benchmark datasets fully validated the effectiveness of our proposed EFV++ framework. New state-of-the-art performance is achieved on a public event stream-based benchmark dataset, i.e., $90.51\%$ on the Bullying10k dataset~\cite{dong2024bullying10k}.

This paper is an extension of our previous work~\cite{yuan2023EFV} which was published at the $6^{th}$ Chinese Conference on Pattern Recognition and Computer Vision (PRCV-2023)\footnote{\url{https://www.prcv2023.cn/2023PRCV}}. 
The main changes can be summarized as follows: 
\textit{1). More Advanced Design at the Feature Interaction Level}: 
In our previous work EFV~\cite{yuan2023EFV}, we directly fuse the dual branches using a bottleneck Transformer network. Such a fusion scheme achieves good performance but ignores the quality of the features which is very important for multi-view/-modality learning. In this work, we propose a novel differentiated framework in which the high-quality features of each branch are well maintained, and the medium and low features are fused and replaced. 
\textit{2). Novel Feature Readout Mechanism}:
Existing multi-view/-modality learning frameworks usually directly fuse the enhanced features, but ignore the order relations. We have developed a feature readout scheme based on GRU and found that it indeed improves the final recognition performance. 
\textit{3). More Complete Experimental Verification}: 
In addition to the three datasets (i.e., ASL-DVS~\cite{bi2020graph}, N-Caltech101~\cite{he2015spatial}, N-MNIST~\cite{orchard2015converting}) evaluated in our conference paper, we also report and compare our model with other state-of-the-art algorithms on the recently released large-scale event stream-based recognition datasets PokerEvent~\cite{wang2023sstformer} and Bullying10k~\cite{dong2024bullying10k}. More in-depth experiments are also conducted to help the readers better understand our newly proposed EFV++ framework.

\textit{The following of this work is organized as}: 
We give an introduction to the related works in section~\ref{sec:relatedworks}. The EFV++ framework we proposed is mainly described in section~\ref{sec:Method}, with a focus on the overview, input representation, each sub-network, classification head, and loss function. We conduct extensive experiments in section~\ref{sec:Experiments} and mainly introduce the datasets and evaluation metrics, implementation details, comparisons with other SOTA models, ablation study, qualitative analysis, and limitation analysis. Finally, we conclude this paper and propose possible improvements as our future works in section~\ref{sec:conclusions}.

\section{Related Works}  \label{sec:relatedworks} 
 In this section, we give an introduction to Event stream-based Recognition\footnote{\url{https://github.com/Event-AHU/Event_Camera_in_Top_Conference}}, Multi-View/-Modal based Recognition, and Transformer Network.

\subsection{Event Stream-based Recognition}  
Current research on event stream-based recognition can be divided into four distinct aspects:  
CNN (Convolutional Neural Networks)-based~\cite{wang2019ev}, 
SNN (Spiking Neural Networks)-based~\cite{fang2020exploiting, fang2021incorporating}, 
GNN (Graph Neural Networks)-based~\cite{bi2019graph, bi2020graph, wang2021event}, and Transformer based models~\cite{wang2022hardvs}. 
For the CNN-based models, Wang et al.~\cite{wang2019ev} proposed an event stream-based gait recognition (EV-gait) method, which effectively removes noise via motion consistency. 
SNN is also utilized for encoding the event stream to achieve energy-efficient recognition. A highly efficient conversion of ANN (Artificial Neural Network) to SNN method is put forward by Peter et al.~\cite{diehl2015fast}, the method involves the balance of the weights and thresholds, while achieving lower latency and requiring fewer operations. 
In~\cite{perez2021sparse}, a sparse back-propagation method for SNN was introduced by redefining the surrogate gradient function form. Fang et al.\cite{fang2021deep} propose spike element-wise (SEW) ResNet to implement residual learning for deep SNNs, while proving that SEW ResNet can easily implement identity mapping and overcome the gradient vanishing/explosion problem of Spiking ResNet. Meng et al.~\cite{meng2022training} propose an accurate and low latency SNN based on the Differentiation on Spike Representation (DSR) method.

For the point cloud based representation, Wang et al.~\cite{wang2019space} treat the event stream as space-time event clouds and adopt PointNet~\cite{qi2017pointnet} as their backbone for gesture recognition. 
Sai et al. propose the event variational auto-encoder (eVAE)~\cite{vemprala2021representation} to directly achieve compact representation learning from the asynchronous event points. 
VMV-GCN~\cite{xie2022vmv} proposed by Xie et al. is a voxel-wise graph learning model to fuse multi-view volumetric. 
Different from these works, in this work, we propose to use both image and voxel to represent the event and then fuse them hierarchically based on the feature quality. Also, our newly proposed readout mechanism works better than a simple concatenate operation for multi-view event recognition.

\subsection{Multi-View/-Modality based Recognition} 
Fusing multi-view/-modality cues is an effective way to achieve more robust and accurate performance. 
To be specific, Wang et al. propose a hybrid SNN-ANN framework for RGB-Event-based recognition by fusing the memory support Transformer and spiking neural networks, termed SSTFormer~\cite{wang2023sstformer}. Deng et al. propose \cite{deng2024dyGCNEventRepr} cross-representation distillation (CRD) framework which transfers prior knowledge from the dense events representation for the event graph-based recognition. Wang et al. propose to track the target objects using a hierarchical cross-modality distillation framework~\cite{wang2023eventvot}. Li et al. propose the SAFE~\cite{li2023SAFE} for RGB-Event based pattern recognition by fusing the RGB, Event frames, and semantic labels using self-attention and cross-attention. 
To more accurately locate regions and better model cross-/intra-modal relationships, Jia et al. propose a machine reading comprehension-based MNER framework called MRC-MNER~\cite{jia2022query}. Bruce et al. propose a model-based multimodal network (MMNet)~\cite{bruce2022mmnet} that uses spatiotemporal graph convolutional networks to learn skeleton modalities to learn attention weights, which will be transferred to the network of RGB modalities. A multi-view attention-aware network is proposed by Deng et al.~\cite{deng2021mvfnet} which projects event stream into multi-view 2D maps to use off-the-shelf 2D models and spatiotemporal complements for recognition. Inspired by the success of the multi-view or multi-modal works, we design a spatial-stereo fusion framework for high-performance event stream recognition.

\subsection{Transformer Network}    
The Transformer proposed in the field of natural language processing achieved excellent performance~\cite{vaswani2017attention} and is also adapted in other research communities, e.g., computer vision, and multi-modal~\cite{wang2023MMPTMs}. 
To be specific, Vision Transformer (ViT)~\cite{dosovitskiy2020image} partitions the image into a series of non-overlapping patches and projects them into tokens as the input of self-attention modules. Their success is owed to the long-term relationship mining between various tokens. 
DeiT~\cite{touvron2021going} proposes a strategy based on token distillation using convolutional networks as the teacher network, which addresses the issue of requiring a large amount of data for pre-training. 
ConViT~\cite{d2021convit} introduces the inductive bias of CNN into the Transformer and achieves better results. VOLO~\cite{yuan2022volo} uses a two-stage structure, which can generate attention weights corresponding to its surrounding tokens through simple linear transformations, avoiding the expensive computational cost of the original self-attention mechanism. 
Swin Transformer~\cite{liu2021swin} introduces a sliding window mechanism, enabling the model to handle super-resolution images, and solving the problem of high computational complexity in the ViT model. 
MoCo v3~\cite{chen2021empirical} applies contrastive learning for ViT-based self-supervised learning. 
He et al. randomly mask a high-proportion of patches under the auto-encoder framework and propose the MAE~\cite{he2022masked} for self-supervised learning. 
Pre-trained multi-modal big models like the CLIP~\cite{radford2021learning} are also developed based on vision Transformer backbone networks. 
In this work, we also adopt the Transformer backbone networks to learn the spatial-temporal features of event images and fuse them with features output from the GNN sub-network.

\begin{figure*}
\centering
\includegraphics[width=\linewidth]{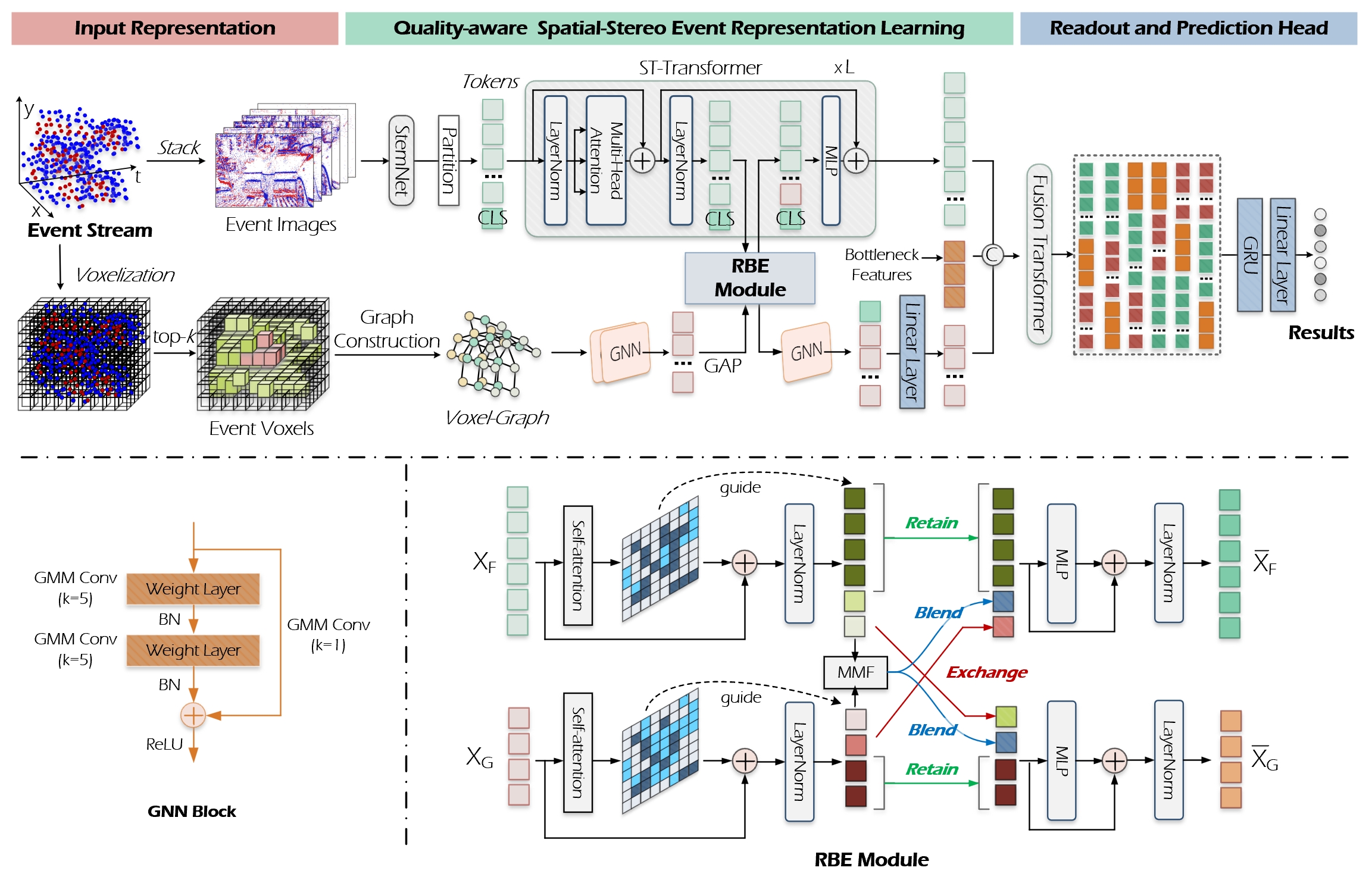}
\caption{
An illustration of our proposed EFV++ framework which takes the diverse event stream representations as the input, i.e., the event frame and voxel. We adopt the Transformer and graph neural networks to handle the event frames and event voxels respectively, and fuse the dual stream in a differentiable manner. To be specific, we keep high-quality features, remove redundant and ineffective features, and integrate general features. Then, we combine the output features from the two branches with bottleneck features, arrange them, and then use a GRU network for fusion to obtain a more diverse feature representation. Finally, we feed this feature into the classification head for effective pattern recognition. 
}
\label{fig:framework}
\end{figure*}

\section{Our Proposed Approach}  \label{sec:Method} 

This section will first provide an overview of our proposed framework. Then, we focus on introducing the \textit{Input Representation}, \textit{Transformer and GNN Backbone Networks}, \textit{Quality-aware RBE Module}, \textit{Hybrid Interaction Readout Mechanism}, and \textit{Loss Function}.

\subsection{Overview}  
Given the event stream, we first transform them into event frames and event voxels to get diverse event representations. For the event image, we adopt a stem network to embed the frames into feature maps and partition them into event tokens. For the event voxel, we select the top-$k$ voxels as the node and connect the nearby nodes using edges to build a voxel graph. Then, the spatial-temporal Transformer networks and graph neural networks are adopted to learn the features of each view. More importantly, we design a novel quality-aware RBE module that achieves exchange-based information interactive learning between the dual features by considering the quality of features. Specifically, we retain the effective features, discard the redundant features, and organically fuse the regular features. After that, we introduce a bottleneck Transformer to fuse these features. We also consider different feature permutations and propose a new readout mechanism to achieve a higher recognition performance. More details of the key modules and procedures will be introduced in the subsequent sub-sections respectively.

\subsection{Input Representation} 
Due to the dense temporal nature of event stream data, directly processing long event streams would entail significant computational overhead. Therefore, existing algorithms typically employ down-sampling methods to sparsify the data. This paper follows a similar approach, transforming dense asynchronous event streams into coarser voxel and synchronous event frame representations to retain as much of the original spatio-temporal information as possible. Assuming the raw event stream is denoted as $\mathcal{E} \in \mathbb{R}^{H \times W \times T}$, we first segment it into event voxels $\mathcal{E}_v \in \mathbb{R}^{h \times w \times t}$ and select the top-$K$ voxels to retain a more complete spatio-temporal representation by eliminating event voxels with no informational content. Let $\mathcal{O}=\{o_1, o_2, \cdots, o_K\}$ represents the set of the ultimately selected voxels, where each event voxel $o_i$ is linked with a feature descriptor $a_i \in \mathbb{R}^C$ that encapsulates its associated events' attributes (polarity). Therefore, each $o_i\in \mathcal{O}$ is depicted as $o_i=(x_i, y_i, t_i, a_i)$, wherein $x_i, y_i, t_i$ refers to the 3D coordinate of each voxel. In addition, we attempt to combine different event stream representations to better leverage the advantages of their diverse representations. Thus, we have the \textit{event frames} and \textit{event voxels} as the input of our network.

\subsection{Backbone Networks}  
For the obtained event voxels $\mathcal{O}$, we construct a geometric neighboring graph $\mathcal{G} (V^o, E^o)$, with each node $v_i \in V^o$ representing a voxel $o_i = (x_i, y_i, t_i, a_i) \in \mathcal{O}$, described as a feature vector $a_i \in \mathbb{R}^{C}$. An edge $e_{ij} \in E^o$ exists between nodes $v_i$ and $v_j$ if their 3D coordinates' Euclidean distance is below a threshold $R$. Inspired by~\cite{wang2021event}, we apply the Gaussian Mixture Model (GMM)-based graph convolution to learn effective voxel graph representations. Each event node $v_i$ aggregates features from its adjacency nodes in each GCN (Graph Convolutional Network) layer in this procedure. Then, we use the average graph pooling to gain the global representation of the voxel graph. A detailed GNN block is shown in the bottom left of Fig.~\ref{fig:framework} and we prefer the readers to check \cite{wang2021event} for more details about the GMM-based convolution network.

Meanwhile, we process the event frames to realize high-quality spatial feature learning, whose dimension is $T \times H \times W \times 3$, where $T$ denotes the number of frames in each event sequence and $H, W$ is the scale of each event frame. A stem network (ResNet18~\cite{he2016deep} in our experiments) is adopted to embed the raw event frames into feature tokens. Then, we employ a spatial-temporal Transformer (ST-Transformer for short) for global feature learning. It mainly contains Multi-head Self-Attention (MSA), Multi-Layer Perceptron (MLP), and Layer Norm (LN) operators. More in detail, we divide each event frame into $N$ spatial patches and get $T \times N$ visual tokens. The learnable position encoding is also considered by adding the visual tokens to maintain their spatial coordinate information.

Based on the aforementioned features, existing works usually directly fuse the features of the event frame and event voxel using a multi-modal fusion module~\cite{wang2023sstformer}. However, we believe that features within the same modality may have representations of varying quality. A brute-force fusion of these features may not yield optimal results. Therefore, this paper proposes a discriminative treatment of features, retaining high-quality features to preserve the unique modality-specific representations, directly replacing low-quality features, and fusing moderate-quality features to obtain a more robust feature representation. We will focus on the RBE (Retain, Blend, and Exchange) module in the following sub-section.

\subsection{Quality-aware RBE Module} 
\begin{figure*}
\centering
\includegraphics[width=0.95\textwidth]{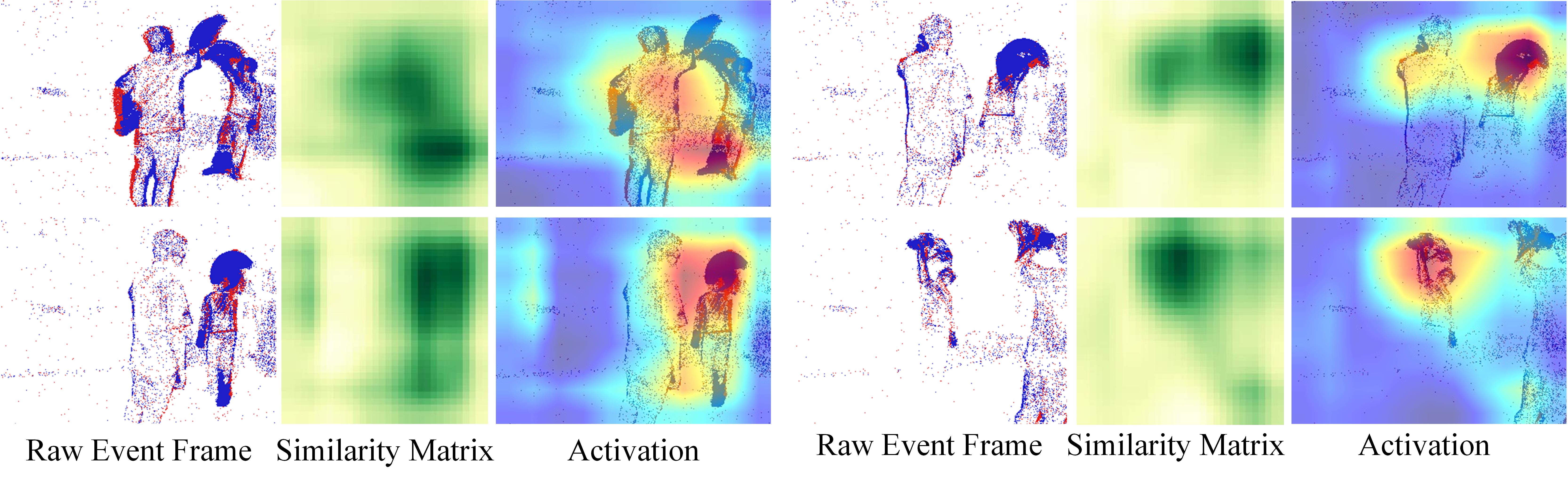}
\caption{   Visualization of similarity matrix learned by ST-Transformer in the RBE module. 
    The $2^{th}$ and $5^{th}$ columns are raw attention matrix, and the $3^{th}$ and $6^{th}$ columns are superimposed images of the input image and the attention map after resizing. }
\label{fig:similarity_VIS}
\end{figure*}

Based on the aforementioned analysis, in this sub-section, we introduce the quality-aware RBE (Retain, Blend, and Exchange) module further to enhance the feature representations of the backbone network. Assume the features obtained from the ST-Transformer sub-network and GNN sub-network are denoted as $\textbf{X}_F$ and $\textbf{X}_G$, respectively. We feed both of them into self-attention layers to get the attention matrices $\textbf{W}_F$, $\textbf{W}_G$, and layer normalization to produce enhanced features. Next, the attention matrices are converted into CLS weights, denoted as $CLS_F$ and $CLS_G$, by summarizing each row of the matrices into a real value. As shown in Fig.~\ref{fig:similarity_VIS}, we can find that the attention matrix reflects the relations between different tokens in the event frames well. Therefore, this dynamically learnable attention is a good indicator for measuring the merits of different features.

For each branch, we introduce two thresholds $\theta_1$ and $\theta_2$ to separate the features into high-, medium-, and low-quality parts. Specifically, we retain the high-quality features (the feature whose attention is larger than $\theta_1$) for each branch. For the medium features (attention weights range from [$\theta_1$, $\theta_2$]), we fuse them with another branch to further enhance its representation. For the low-quality features whose attention weights are less than $\theta_2$, we directly replace them using average features from the other branch. Thus, the processed features are denoted as $\Bar{\textbf{X}}_F$ and $\Bar{\textbf{X}}_G$.

\subsection{Hybrid Interaction Readout Mechanism} 

To further facilitate the interactions between the dual views, we introduce bottleneck tokens $\textbf{X}_B$ to connect them by feeding them into the Transformer layer. 
\begin{equation}
    \label{bottleneckTransformer} 
    [\mathcal{X}_F, \mathcal{X}_G, \mathcal{X}_B] = Transformer([\Bar{\textbf{X}}_F, \Bar{\textbf{X}}_G, \textbf{X}_B])
\end{equation}
where [, ] denotes the concatenate operation. The outputs are denoted as $\mathcal{X}_F$, $\mathcal{X}_G$, and $\mathcal{X}_B$. Usually, we can concatenate these three features into one representation for recognition. However, how to arrange these sets of features, or whether different arrangements can lead to diverse results, is an interesting issue worth exploring.

In this paper, we are the first to attempt arranging and combining these features and then fusing them with a GRU network, resulting in a richer feature representation that further enhances the recognition results. More in detail, we first obtain a diverse set of expressions $T_1$-$T_6$ through different permutations and combinations, as follows: 
\begin{align*}
T_1 &= \{\mathcal{X}_F, \mathcal{X}_G, \mathcal{X}_B\}, &T_2 &= \{\mathcal{X}_F, \mathcal{X}_B, \mathcal{X}_G\}, \\
T_3 &= \{\mathcal{X}_G, \mathcal{X}_F, \mathcal{X}_B\}, &T_4 &= \{\mathcal{X}_G, \mathcal{X}_B, \mathcal{X}_F\}, \\
T_5 &= \{\mathcal{X}_B, \mathcal{X}_F, \mathcal{X}_G\}, &T_6 &= \{\mathcal{X}_B, \mathcal{X}_G, \mathcal{X}_F\}. \\ 
\end{align*}
Then, as shown in Fig.~\ref{fig:framework}, we adopt the GRU (Gated Recurrent Unit) to fuse the aforementioned features into one unified representation:  
\begin{equation}
h_i = GRU(T_i, h_{i-1}), ~~~i \in \{1, 2, ..., 6\}, 
\end{equation}
where $h_i$ denotes the hidden state of GRU network. In our experiments, we select the hidden state output in the final timestep as enhanced features for recognition.

We believe that the \textit{possible reasons} for the effectiveness of the hybrid interaction readout mechanism are as follows: 
\textit{1). Increased Model Capacity}: The dimensionality of the input features is increased by combining different features and increases the model’s capacity effectively. Thus, more complex functions can be learned to fit the data well. 
\textit{2). Enhanced Interactions}: We believe the interactions between features can be revealed well using the different combinations. It helps in mastering more complex patterns and relationships.




\subsection{Classification Head \& Loss Function} 
After getting the features from the hybrid interaction readout mechanism, we feed them into the classification head which is a two-layer Multi-layer Perceptron (MLP) for recognition. We utilize the \textit{Negative Log Likelihood Loss function} for the training of our framework which measures the difference between the predicted probability distribution of the model and the true label distribution.

\section{Experiments}  \label{sec:Experiments} 

\subsection{Datasets and Evaluation Metrics} \label{datasetsMetric}
In this work, we employ five datasets to evaluate our proposed model, including \textbf{N-caltech101}~\cite{he2015spatial}, \textbf{N-MNIST}~\cite{orchard2015converting}, \textbf{ASL-DVS}~\cite{bi2020graph}, \textbf{Bullying10k}~\cite{dong2024bullying10k} and \textbf{PokerEvent} dataset~\cite{wang2023sstformer}. A brief introduction to these datasets is given below: 

\noindent $\bullet$ \textbf{ASL-DVS}~\cite{bi2020graph}\footnote{\url{https://github.com/PIX2NVS/NVS2Graph}}: This dataset consists of 100,800 samples, with 4,200 samples available for each letter. The focus was on the 24 letters representing the handshapes of American Sign Language. Each video in this dataset has a duration of approximately 100 milliseconds. The author captured these samples using an iniLabs DAVIS240c camera under realistic conditions. 

\noindent $\bullet$ \textbf{N-Caltech101}~\cite{he2015spatial}\footnote{\url{https://www.garrickorchard.com/datasets/n-caltech101}}: This dataset is short for Neuromorphic-Caltech101, which is obtained by transforming the frame-based Caltech101 dataset into a spike version. It contains 101 types of images, with approximately 40-800 images per class, most of which are 50 images per class. The ATIS sensor is adopted for the sample collection by recording the Caltech101 examples on an LCD monitor. The training and testing subset contains 6932 and 1777 samples, respectively.

\noindent $\bullet$ \textbf{N-MNIST}~\cite{orchard2015converting}\footnote{\url{https://www.garrickorchard.com/datasets/n-mnist}}: This dataset is obtained by recording the display equipment when visualizing the original MNIST (28 $\times$ 28 pixels). The ATIS event camera is used for the data collection and each event sample lasts about 10ms. There are 70,000 event files for this dataset, the training and testing subset contains 60,000 and 10,000 videos, respectively.

\noindent $\bullet$ \textbf{Bullying10k}~\cite{dong2024bullying10k}\footnote{\url{https://www.brain-cog.network/dataset/Bullying10k/}}: This dataset is a privacy-preserving dataset designed for bullying detection. It was captured using a DVS (Dynamic Vision Sensor) camera and includes 10,000 samples, comprising a total of 12 billion events. The training and validation subset contains 7874 and 1970 samples, respectively.

\noindent $\bullet$ \textbf{PokerEvent}~\cite{wang2023sstformer}\footnote{\url{https://github.com/Event-AHU/SSTFormer}}: This dataset focuses on recognizing character patterns in poker cards. It consists of 114 classes and includes 24,415 RGB-Event samples recorded using a DVS346 event camera. The dataset is divided into training and testing subsets, with 16,216 and 8,199 samples, respectively. 

Note that the \textit{top-1 accuracy} is employed as the evaluation metric throughout our study.

\subsection{Implementation Details}   
Our proposed dual-stream event-based recognition framework can be trained in an end-to-end manner. In the training phase, we set the batch size to 8 and train the model for a total of 150 epochs. The initial learning rate is set as 0.001 and multiplied by 0.1 for every 80 epochs. AdamW optimizer~\cite{loshchilov2017decoupled} is adopted for the training of our network. We select eight frames for each video sample and divide each frame into eight tokens. For the constructed voxel graph, the threshold \textit{R} is set to 2. The scale of the voxel grid is (4, 4, 4) for the ASL-DVS dataset. We select 512 voxels as the graph node for the structured graph representation learning. For the Bullying10k dataset, we conduct our experiments on a server with NVIDIA GeForce RTX 3090 GPU which takes approximately 10 hours to complete. The training on the PokerEvent dataset takes about 50 hours using two NVIDIA GeForce RTX 3090 GPUs. More details can be found in our source code.

\subsection{Comparison on Public Benchmarks}

$\bullet$ \textbf{Results on ASL-DVS dataset}~\cite{bi2020graph}
As shown in Table~\ref{ASLDVSResults}, we report our experimental results and compare them with other state-of-the-art algorithms on the ASL-DVS dataset. One can find that our proposed EFV++ achieves the best accuracy on the ASL-DVS dataset, reaching 0.999. Compared with other models, our results exceed theirs by a large margin which indicates superior performance in event stream-based pattern recognition. 

\begin{table}[!htp]
\center   
\caption{Results on the ASL-DVS~\cite{bi2020graph} dataset.} 
\label{ASLDVSResults}
\resizebox{\columnwidth}{!}{ 
\begin{tabular}{ccccccccccccccc} 		
\hline \toprule [0.5 pt] 
\textbf{EST}\cite{gehrig2019end}   &\textbf{AMAE}\cite{deng2020amae}     &\textbf{M-LSTM}\cite{cannici2020differentiable}    &\textbf{MVF-Net}\cite{deng2021mvf}     & \textbf{EventNet}\cite{sekikawa2019eventnet}\\  
0.979   & 0.984     &0.980     &0.971     &0.833    \\ 
\hline 
\textbf{RG-CNNs}\cite{bi2020graph}     &\textbf{\makecell[c]{EV-VGCNN}}\cite{deng2021evvgcnn}     &\textbf{\makecell[c]{VMV-GCN}}\cite{xie2022vmv}     &\textbf{EV-Gait-3DGraph}\cite{wang2019ev}   &\textbf{Ours} \\
0.901     &0.983     &0.989  &0.738  &0.999	 \\
\hline \toprule [0.5 pt] 
\end{tabular}
}
\end{table}

$\bullet$ \textbf{Results on N-Caltech101 dataset}~\cite{he2015spatial} 
As shown in Table~\ref{NcaltechResults}, our model achieves 0.897 on the top-1 accuracy on the N-Caltech101 dataset, meanwhile, other SOTA models obtain inferior performance significantly. The second-ranked model is the VMV-GCN~\cite{xie2022vmv} which achieves 0.778 on this dataset but is still worse than ours. Note that, the VMV-GCN is a voxel-wise graph learning model for spatiotemporal feature learning on the event stream. Therefore, we can draw the conclusion that our model performs well on both event feature learning and fusion for high-performance event stream recognition.

\begin{table}[!htp]
\center   
\caption{Results on the N-Caltech101~\cite{he2015spatial} dataset.} 
\label{NcaltechResults}
\resizebox{\columnwidth}{!}{ 
\begin{tabular}{ccccccccccccccc} 		
\hline \toprule [0.5 pt] 
\textbf{EventNet}\cite{sekikawa2019eventnet}   &\textbf{Gabor-SNN}\cite{sironi2018hats}     &\textbf{RG-CNNs}\cite{bi2020graph}    &\textbf{VMV-GCN}\cite{xie2022vmv}     & \textbf{EV-VGCNN}\cite{deng2021evvgcnn}    &\textbf{EST}\cite{gehrig2019end}\\  
0.425   & 0.196     &0.657     &0.778     &0.748  &0.753  \\ 
\hline 
\textbf{ResNet-50}\cite{He_2016_CVPR}     &\textbf{\makecell[c]{MVF-Net}}\cite{deng2021mvf}     &\textbf{\makecell[c]{M-LSTM}}\cite{cannici2020differentiable}     &\textbf{AMAE}\cite{deng2020amae}    &\textbf{HATS}\cite{sironi2018hats}  &\textbf{Ours} \\
0.637     &0.687     &0.738  &0.694  &0.642	 &0.897 \\
\hline \toprule [0.5 pt] 
\end{tabular}
}
\end{table}

$\bullet$ \textbf{Results on N-MNIST dataset}~\cite{orchard2015converting} 
As illustrated in Table~\ref{NMNISTResults}, many SOTA models achieve superior performance on this benchmark dataset. For example, the EST~\cite{gehrig2019end} and M-LSTM~\cite{cannici2020differentiable} achieve 0.990 and 0.986, RG-CNNs~\cite{bi2020graph} and EV-VGCNN~\cite{deng2021evvgcnn} also obtain 0.990 and 0.994. In contrast, our model also achieves 0.990 on this dataset which is comparable to these SOTA algorithms, which fully demonstrate the effectiveness of our model.

\begin{table}[!htp]
\center   
\caption{Results on the N-MNIST~\cite{orchard2015converting} dataset.} 
\label{NMNISTResults}
\resizebox{\columnwidth}{!}{ 
\begin{tabular}{ccccccccccccccc} 		
\hline \toprule [0.5 pt] 
\textbf{EST}\cite{gehrig2019end}   &\textbf{M-LSTM}\cite{cannici2020differentiable}     &\textbf{Gabor-SNN}\cite{sironi2018hats}    &\textbf{MVF-Net}\cite{deng2021mvf}     & \textbf{EvS-S}\cite{li2021graph}\\  
0.990    &0.986    &0.837    &0.981    &0.991    \\ 
\hline 
\textbf{RG-CNNs}\cite{bi2020graph}     &\textbf{\makecell[c]{EV-VGCNN}}\cite{deng2021evvgcnn}     &\textbf{\makecell[c]{HATS}}\cite{sironi2018hats}     &\textbf{EventNet}\cite{sekikawa2019eventnet}   &\textbf{Ours} \\
0.990    &0.994    &0.991    &0.752    &0.990    \\
\hline \toprule [0.5 pt] 
\end{tabular}
}
\end{table}

$\bullet$ \textbf{Results on Bullying10k dataset}~\cite{dong2024bullying10k} 
In addition to the three datasets mentioned above, which are nearly saturated in terms of performance, we have also conducted experiments on the latest proposed large-scale Bullying10k dataset. As shown in Table~\ref{Bullying10k_acc}, we can find that our proposed model achieves 90.51\% on this dataset, while other strong baselines are significantly inferior to ours. To be specific, SimpleBaseline~\cite{xiao2018simple} and HRNet~\cite{sun2019deep} are the only two models that exceed 88\% on the top-1 accuracy metric. These experiments fully validated the effectiveness of our RBE-based spatial-stereo fusion for event recognition.

\begin{table}[!htp]
\center
\small
\caption{Results on the Bullying10k~\cite{dong2024bullying10k} dataset.} 
\label{Bullying10k_acc}
\resizebox{\columnwidth}{!}{ 
\begin{tabular}{l|c|c|c}
\hline \toprule [0.5 pt]  
\textbf{Algorithm} & \textbf{Source}      &\textbf{Backbone}  &\textbf{Results}  \\
\hline
\textbf{C3D} \cite{ji20123d} & ICCV-2015     &3D-CNN   &71.25     \\
\textbf{R2Plus1D} \cite{tran2018closer} & CVPR-2018   &ResNet-18   &69.25  \\
\textbf{R3D} \cite{tran2017convnet} &arXiv-2017   &ResNet-18   &72.50    \\
\textbf{TAM} \cite{liu2021tam} &  ICCV-2021    &ResNet-50    &71.20     \\
\textbf{SlowFast} \cite{feichtenhofer2019slowfast} & ICCV-2019      &ResNet-50   &74.00  \\ 
\textbf{SNN} \cite{fang2021deep}  &   NeurIPS-2021    &SEW-ResNet19  &67.05   \\ 
\textbf{X3D} \cite{feichtenhofer2020x3d}  &  CVPR-2020       &ResNet   &76.90    \\
\textbf{SimpleBaseline} \cite{xiao2018simple}  &    ECCV-2018     &ResNet-50   &88.30     \\
\textbf{HRNet} \cite{sun2019deep}  &   CVPR-2019      &HRNet-ws32   &88.20     \\
\hline
\textbf{Ours}      &--  &Former-GNN   &90.51    \\ 
\hline \toprule [0.5 pt]  
\end{tabular}}
\end{table}

$\bullet$ \textbf{Results on PokerEvent dataset}~\cite{wang2023sstformer} 
On the PokerEvent dataset, our model achieves 55.40\% which is even competitive to other multi-modal algorithms, such as the TSM~\cite{lin2019tsm} and TimeSformer~\cite{bertasius2021TimeSformer}. Our results are also better than most of the compared algorithms, including the recently proposed memory-support Transformer (MST)~\cite{wang2023sstformer}, MVIT~\cite{li2022mvitv2}, X3D~\cite{feichtenhofer2020x3d} which all takes the RGB and event data as their input on this benchmark. These comparisons fully demonstrate the effectiveness of our proposed modules for event stream recognition.

\begin{table}[!htp]
\scriptsize         
\center 
\caption{Results on the PokerEvent dataset. V+E denotes the input data are visible frames and event stream, otherwise, event only.}    
\label{pokerResults}
\resizebox{0.48\textwidth}{!}{
\begin{tabular}{c|l|c|cccc} 
\hline \toprule [0.5 pt]
\textbf{Algorithm} &\textbf{Publish} &\textbf{V+E} &\textbf{Backbone} &\textbf{Results}    \\
\hline 
\textbf{C3D}~\cite{ji20123d} &ICCV-2015  &\cmark    &3D-CNN   &51.76              \\
\textbf{TSM}~\cite{lin2019tsm} &ICCV-2019  &\cmark  &ResNet-50       &55.43     \\
\textbf{ACTION-Net}~\cite{wang2021actionnet}  &CVPR-2021  &\cmark       &ResNet-50   &54.29 \\
\textbf{TAM}~\cite{liu2021tam} 	&ICCV-2021     &\cmark        &ResNet-50       &53.65      \\
\textbf{V-SwinTrans}~\cite{liu2022videoswintransformer}   &CVPR-2022     &\cmark        &Swin-Former      &54.17          \\
\textbf{TimeSformer}~\cite{bertasius2021TimeSformer}  &ICML-2021     &\cmark        &ViT      &55.69        \\
\textbf{X3D}~\cite{feichtenhofer2020x3d}  &CVPR-2020     &\cmark        &ResNet       & 51.75           \\
\textbf{MVIT}~\cite{li2022mvitv2}  &CVPR-2022   &\cmark          &ViT       &55.02       \\
\textbf{SCNN-MST}~\cite{wang2023sstformer}  &arXiv2024    &\cmark         &SNN-Former       &53.19       \\  
\textbf{Spikingformer-MST}~\cite{wang2023sstformer}  &arXiv2024     &\cmark        &SNN-Former       &54.74       \\
\hline 
\textbf{Ours (Event Only)}  &-    &\xmark         &Former-GNN       &55.40       \\
\hline \toprule [0.5 pt]
\end{tabular} }   
\end{table}

\begin{table*}[!htp]
\centering
\caption{Ablation study on N-Caltech101, ASL-DVS, and PokerEvent datasets.} 
\label{AblationResults}
\begin{tabular}{l|l|l|l|l} 		  
\hline \toprule [0.5 pt] 
\textbf{Index} & \textbf{Component} & \textbf{N-Caltech101}~\cite{he2015spatial} & \textbf{ASL-DVS}~\cite{bi2020graph} & \textbf{PokerEvent}~\cite{wang2023sstformer} \\  
\hline
\#1 & Baseline (Event Voxel Only)                & 38.89    &   98.73   & 10.87    \\ 
\#2 & Baseline (Event Image Only)                & 86.69     &  98.96    & 53.12     \\ 
\#3 & Baseline (Event Image + Voxel)             & 88.41 & 99.60 & 54.34 \\ 
\hline 
\#4  \scriptsize{(\textit{vs} \#3)} & Baseline (w/o Bottleneck Feature) & 87.11 \scriptsize $\textcolor{DarkRed}{(\downarrow -1.30)}$   &   99.41 \scriptsize $\textcolor{DarkRed}{(\downarrow -0.19)}$   &   53.86 \scriptsize $\textcolor{DarkRed}{(\downarrow -0.48)}$    \\ 
\#5  \scriptsize{(\textit{vs} \#3)} & Baseline (w/o FusionFormer) & 87.56 \scriptsize $\textcolor{DarkRed}{(\downarrow -0.85)}$   & 99.50 \scriptsize $\textcolor{DarkRed}{(\downarrow -0.10)}$     & 54.21 \scriptsize $\textcolor{DarkRed}{(\downarrow -0.13)}$   \\ 
\hline 
\#6 \scriptsize{(\textit{vs} \#3)} & Baseline+RBE             & 89.19 \scriptsize $\textcolor{SeaGreen4}{(\uparrow +0.78)}$  
                            & 99.97 \scriptsize $\textcolor{SeaGreen4}{(\uparrow +0.37)}$ 
                            & 54.86 \scriptsize $\textcolor{SeaGreen4}{(\uparrow +0.52)}$ \\ 
\#7  \scriptsize{(\textit{vs} \#3)} & Baseline+Readout    & 88.52 \scriptsize $\textcolor{SeaGreen4}{(\uparrow +0.11)}$     &   99.96 \scriptsize $\textcolor{SeaGreen4}{(\uparrow +0.36)}$   &54.47 \scriptsize $\textcolor{SeaGreen4}{(\uparrow +0.13)}$      \\
\#8  \scriptsize{(\textit{vs} \#3)} & Baseline+RBE+Readout    & 89.76 \scriptsize $\textcolor{SeaGreen4}{(\uparrow +1.35)}$
                            & 99.99 \scriptsize $\textcolor{SeaGreen4}{(\uparrow +0.39)}$
                            & 55.40 \scriptsize $\textcolor{SeaGreen4}{(\uparrow +1.06)}$ \\ 
\hline \toprule [0.5 pt] 
\end{tabular}
\end{table*}

\begin{figure}[!htp]
    \centering
    \includegraphics[width=1\linewidth]{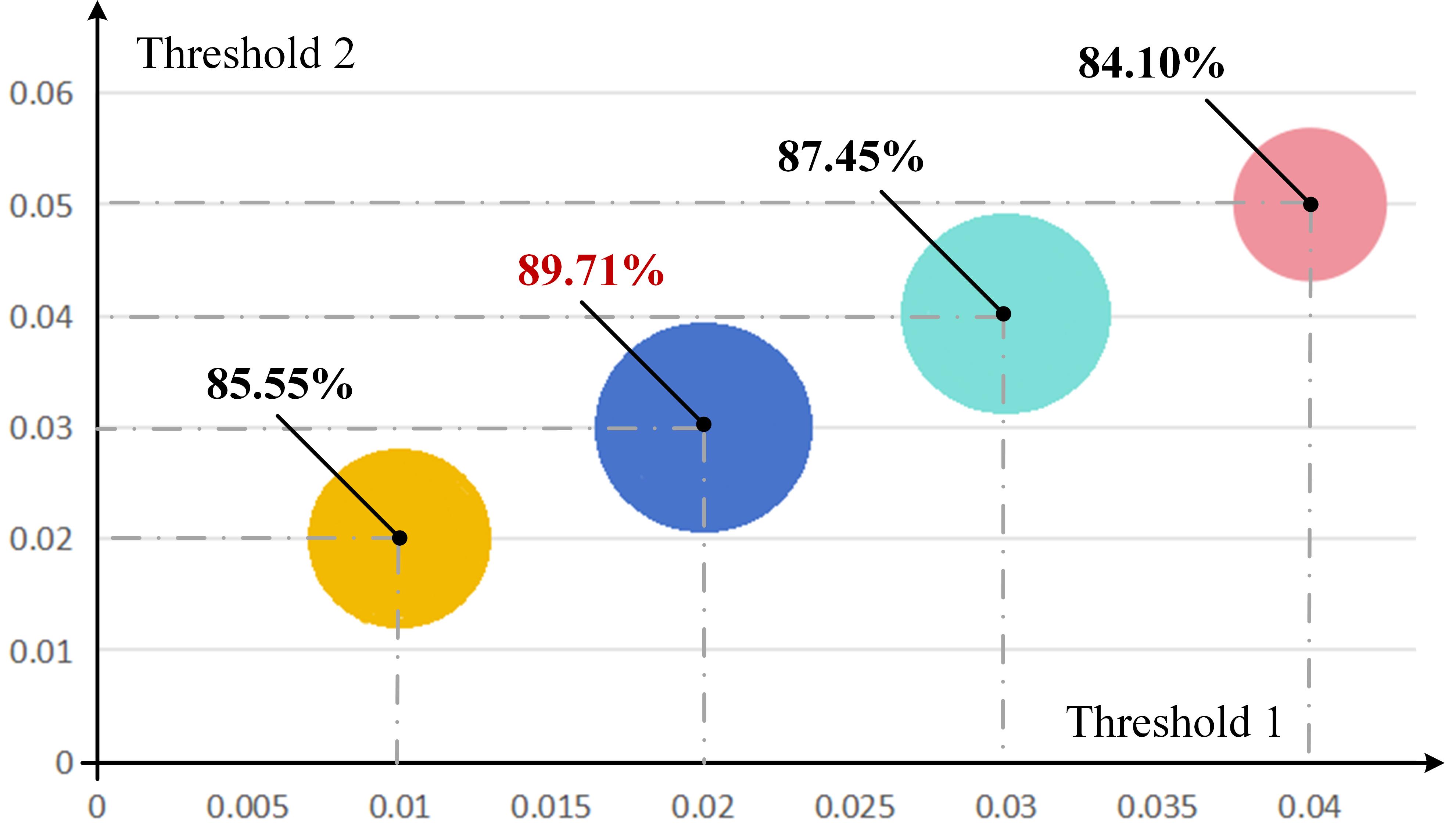}
    \caption{Analysis of key thresholds for recognition on N-Caltech101 dataset.} 
    \label{fig:threshold12Results}
\end{figure}

\subsection{Ablation Study} 

$\bullet$ \textbf{Component Analysis.~} 
To help the readers better understand the effectiveness of the key components in our framework, in this sub-section, we conduct ablation studies to demonstrate their contributions. The performance of our model is evaluated on three datasets, including N-Caltech101, ASL-DVS, and PokerEvent dataset. As shown in Table~\ref{AblationResults}, \textit{Baseline} refers to the basic structure of our model proposed in our conference paper, \textit{RBE} is short for Retain-Blend-Exchange module, and \textit{Readout} is the hybrid interaction readout mechanism we applied to aggregate all the information and generate the final output. 

\begin{figure*}[!htp]
    \centering
    \includegraphics[width=1\textwidth]{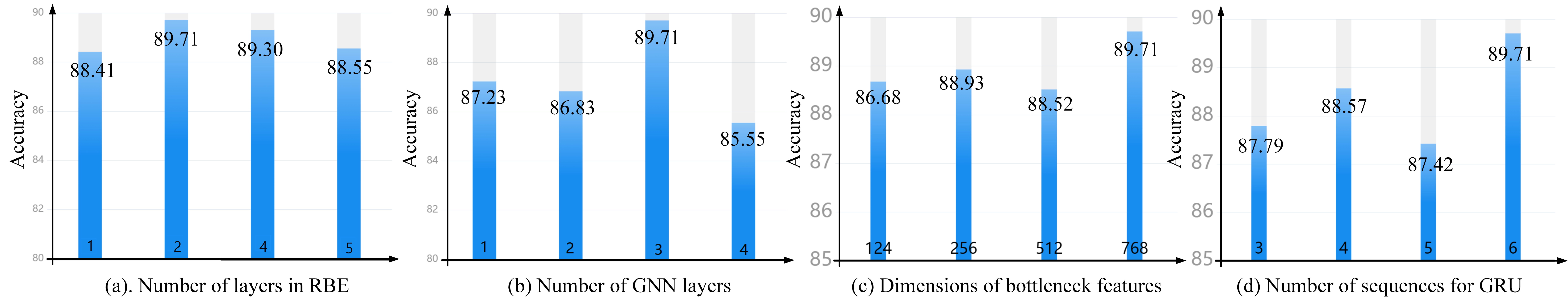}
    \caption{Parameter analysis of core modules in our framework.} 
    \label{fig:paramAnalysis}
\end{figure*}

As shown in Table~\ref{AblationResults}, we test the influence of different event representations for our model, i.e., the event frame and event voxel. In the N-Caltech101 dataset, we can find that when a single representation is used, the results are 86.69\%(event frame) and 38.89\% (event voxel), respectively. When both representations are utilized, the performance can be improved to 88.41\%, which fully validates the effectiveness of dual-view fusion. For the bottleneck Transformer module, we tried to remove the bottleneck features and fusion Transformer, the performance dropped from to 87.11\%, and 87.56\%, respectively. These experiments fully validated the effectiveness of the bottleneck Transformer for our framework. 

For the key modules proposed in this paper, we integrate the RBE and hybrid interaction readout mechanism into the baseline framework, as shown in line 6, 7, and 8 of Table~\ref{AblationResults}. It is easy to find that when introducing the RBE module, the results improved from 88.41, 99.60, and 54.34 to 89.19, 99.97, and 54.86 on the N-Caltech101, ASL-DVS, PokerEvent dataset, respectively. When introducing the new hybrid interaction readout mechanism, these results can be further improved to 89.76, 99.99, and 55.40, respectively, the improvements are up to +1.35, +0.39, +1.06. These results fully demonstrate the effectiveness of our proposed feature interaction operation.

$\bullet$ \textbf{Analysis of the Number of Backbone Layers.~} 
As shown in Fig.~\ref{fig:paramAnalysis} (a), we set different layers of GNN/Transformer for the encoding of event voxel/frame, including 1, 2, 4, and 5. On the N-Caltech101 dataset, we get 88.41, 89.71, 89.30, and 88.55 for these four settings. Therefore, we set 4 four backbone layers as a general setting for other experiments.

$\bullet$ \textbf{Analysis of GNN layers.~} 
As shown in Fig.~\ref{fig:paramAnalysis} (b), we test different GNN layers to encode event voxel, ranging from 1 to 4. From the experimental results, we can find that the impact of the number of layers on the final recognition performance is not linear. When we only have 1 GNN layer, the result is 87.23\%, When increasing it to 2 layers, the results slightly decreased to 86.83\%. These results suggest that adding more layers does not always lead to an improvement. However, when the number of layers is increased to 3, the result surges to an impressive 89.30\%. This is the highest result in the dataset, indicating that three layers might be an optimal configuration for this specific model. Contrarily, when the number of layers is increased to 4, the result dramatically drops to 85.55\%, which is the lowest among the given data. This suggests that beyond a certain point, the increase in layers may have a detrimental effect on the results. 

In conclusion, the number of layers in a GNN significantly impacts the results. The relationship is not linear, which implies the importance of carefully selecting the number of layers. Too many or too few layers can negatively affect the performance, indicating that there exists an optimal number of layers for achieving the best results. According to this dataset, three layers provide the best outcome.

$\bullet$ \textbf{Analysis of Different Dimensions of Bottleneck Features.~} 
In our framework, we introduce the bottleneck Transformer to fuse the event voxels and frames for high-performance recognition. In this part, we test different dimensions for the bottleneck features, i.e., 124, 256, 512, and 768. As illustrated in Fig.~\ref{fig:paramAnalysis} (c), our results are 86.68, 88.93, 88.52, and 89.71, respectively on the N-Caltech101 dataset. We can find that a better outcome can be achieved when 768-D is used.

\begin{figure*}[!htp]
\centering
\includegraphics[width=\textwidth]{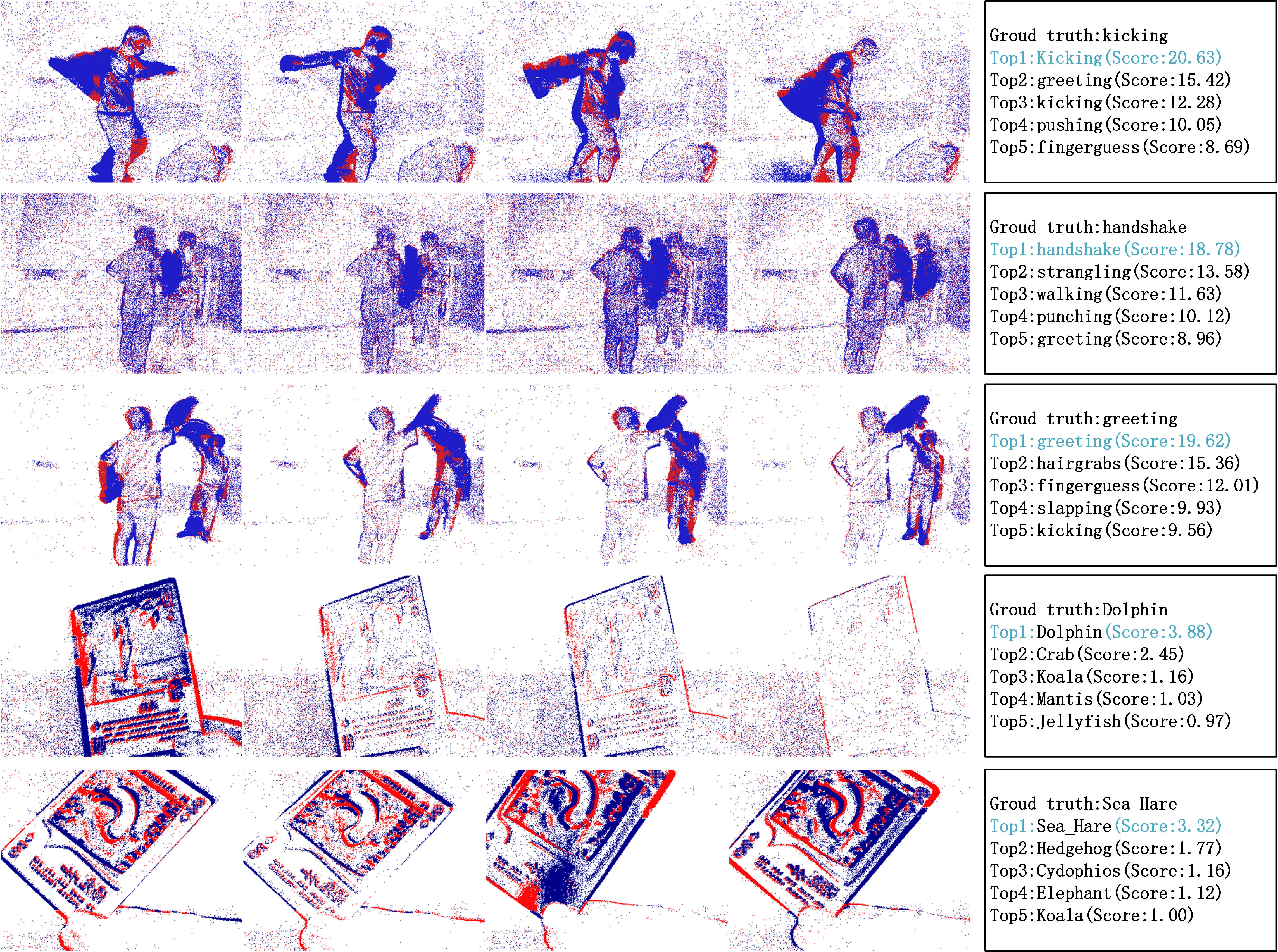}
\caption{The top-5 predictions of our model on the Bully10k (the first 3 rows) and PokerEvent (the bottom 2 rows) datasets.}
\label{fig:top5results}
\end{figure*}

\begin{figure*}[!htp]
\centering
\includegraphics[width=\textwidth]{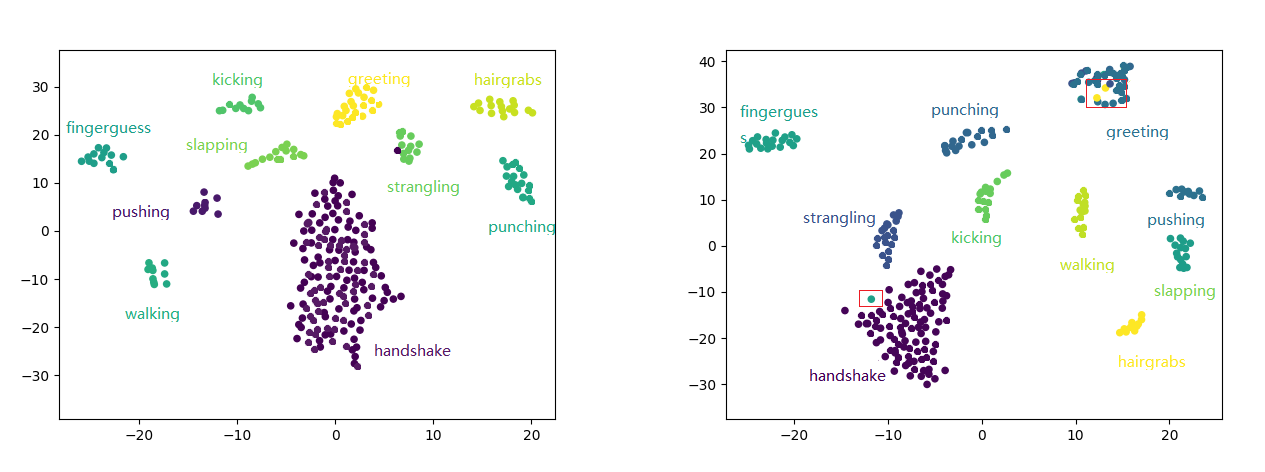}
\caption{ The feature distribution of our newly proposed model (left sub-figure) and baseline~\cite{yuan2023EFV} (right sub-figure) on the Bully10k dataset using T-SNE.}  
\label{fig:tsneresults}
\end{figure*}

$\bullet$ \textbf{Analysis of Different Number of Sequences for Hybrid Interaction Readout Mechanism.~} 
In our experiments, we test all possible combinations for the output feature (in our case, the combinations are 6). As illustrated in Fig.~\ref{fig:paramAnalysis} (d), when we set this parameter as 3, 4, 5, and 6, respectively, the corresponding accuracy is 87.79, 88.57, 87.42, and 89.71. We can find that a better result can be achieved if we utilize all these features for the hybrid interaction readout mechanism.

$\bullet$ \textbf{Analysis of the Threshold in RBE Module.~}
In our framework, the threshold parameters to separate the features into different qualities are rather important. Thus, in this section, we give a visualization of the final recognition performance using different thresholds. As shown in Fig.~\ref{fig:threshold12Results}, when setting threshold 1 and threshold 2 as 0.02 and 0.03, respectively, our model achieves the best results, i.e., 89.71\% on the N-Caltech101 dataset. When changing them into other settings, an inferior result can be obtained. Therefore, we select the optimal threshold parameters when conducting other experiments.

\subsection{Visualization}
In this part, we give a quantitative analysis to enhance the interpretability of our algorithm further. Top-5 recognition results and feature embeddings are provided below, respectively.

$\bullet$ \textbf{Top-5 Recognition Results.~} 
As depicted in Fig.~\ref{fig:top5results}, we display 3 and 2 sets of event samples from PokerEvent and Bullying10k datasets, respectively, alongside their corresponding visualizations of the top five recognition results. These scores represent the response scores for each of the 114 categories in the PokerEvent dataset and 10 categories in the Bullying10k dataset. We can find that our model predicts the right category with a relatively higher response score compared with other classes.

$\bullet$ \textbf{Feature Embedding.~}  
As shown in Fig.~\ref{fig:tsneresults}, we present a compelling visual example that allows us to delve into the distances between different classes. A total of 10 classes are randomly chosen to visualize this feature from the Bully10k dataset. From the comparison between feature embeddings, we can find that our newly proposed strategies further enhance the baseline approach proposed in our conference paper. Note that, the samples in the rectangles are wrongly classified.

\subsection{Limitation Analysis}  
From the experimental results reported above, we can find that our proposed EFV++ performs well on multiple event stream-based pattern recognition datasets. Despite these advances, our framework is still limited by the following issues: 
\textit{1}). We transform the event stream into frames and voxels and attempt to fuse the dual views in a unified framework which makes it challenging for practical employment. Considering the fact that it still provides us with a good performance on sparse event stream recognition, we can treat it as a teacher network and distill a lightweight network for efficient event recognition.  
\textit{2}). This framework makes full use of visual cues well for event stream-based pattern recognition, however, the semantic cues are ignored considering the trade-off between accuracy and complexity. In future works, we will consider introducing semantic categories for a better understanding of the classified patterns.

\section{Conclusion and Future Works} \label{sec:conclusions}  
In this paper, we presented a novel dual-stream framework, EFV++, for event representation, feature extraction, and differentiated fusion. By modeling two common event representations simultaneously, the spatial and three-dimensional stereo information can be learned separately, leading to a more comprehensive understanding of the event stream. Specifically, we adopt the Transformer and Structured GNN to encode the event frames and event stream, respectively. More importantly, we design a novel RBE module that separates the features into high-, medium-, and low-quality ones, and retain-blend-exchange these features to achieve a better fusion. Then, a bottleneck Transformer is adopted to further fuse the dual branches. Finally, we propose a new hybrid interaction readout mechanism to further comprehend the feature representation for pattern recognition. Our experimental results confirm the effectiveness of the EFV++ model, outperforming state-of-the-art approaches on multiple event stream-based classification datasets.

In our future works, we will consider distilling lightweight neural networks based on our proposed EFV++ to make it more hardware-friendly. In addition, the semantic cues are also considered when designing new architecture for the challenging event stream pattern recognition.

{ 
\bibliographystyle{IEEEtran}
\bibliography{reference}
}

\end{document}